\pdfoutput=1

\documentclass[11pt]{article}

\usepackage[]{acl}

\usepackage{times}
\usepackage{latexsym}
\usepackage{graphicx}
\usepackage{amsmath}
\usepackage{multirow}
\usepackage{amssymb}

\usepackage[T1]{fontenc}

\usepackage[utf8]{inputenc}

\usepackage{microtype}

\title{Towards Robust Visual Question Answering: Making the Most of Biased Samples via Contrastive Learning}

\author{Qingyi Si$^{1,2}$,\ Yuanxin Liu$^{1,4}$,\ Fandong Meng$^{3}$, Zheng Lin$^{1,2}$\thanks{\ \  Corresponding author: Zheng Lin.}\  \\  {\bf\ Peng Fu$^{1}$,\ Yanan Cao$^{1,2}$,\ Weiping Wang$^1$,\ Jie Zhou$^3$ } \\ 
$^1$Institute of Information Engineering, Chinese Academy of Sciences, Beijing, China \\
$^2$School of Cyber Security, University of Chinese Academy of Sciences, Beijing, China \\
$^3$Parttern Recognition Center, WeChat AI, Tencent Inc, China \qquad $^4$Peking University \\
  \tt{\small{\{siqingyi,linzheng,fupeng,caoyanan,wangweiping\}@iie.ac.cn,}} \\
  \tt{\small{liuyuanxin@stu.pku.edu.cn}},\tt{\small{\{fandongmeng,withtomzhou\}@tencent.com}}
  }

\begin{document}
\maketitle
\begin{abstract}
Models for Visual Question Answering (VQA) often rely on the spurious correlations, i.e., the language priors, that appear in the biased samples of training set, which make them brittle against the out-of-distribution (OOD) test data. Recent methods have achieved promising progress in overcoming this problem by reducing the impact of biased samples on model training. However, these models reveal a trade-off that the improvements on OOD data severely sacrifice the performance on the in-distribution (ID) data (which is dominated by the biased samples). Therefore, we propose a novel 
contrastive learning approach, MMBS\footnote{Joint work with Pattern Recognition Center, WeChat AI, Tencent Inc, China. The code is available at \url{https://github.com/PhoebusSi/MMBS}.}, for building robust VQA models by \textbf{M}aking the \textbf{M}ost of \textbf{B}iased \textbf{S}amples. Specifically, we construct positive samples for contrastive learning by eliminating the information related to spurious correlation from the original training samples and explore several strategies to use the constructed positive samples for training. 
Instead of undermining the importance of biased samples in model training, our approach precisely exploits the biased samples for unbiased information that contributes to reasoning. The proposed method is compatible with various VQA backbones.  We validate our contributions by achieving competitive performance on the OOD dataset VQA-CP v2 while preserving robust performance on the ID dataset VQA v2. 

\end{abstract}

\section{Introduction}
Visual Question Answering (VQA), aiming to answer a question about the given image, is a multi-modal task that involves the intersection between vision and language. 
Despite the remarkable performance on many VQA datasets such as VQA v2 \citep{goyal2017making}, recent studies \citep{antol2015vqa,kafle2017analysis,agrawal2016analyzing} find that the VQA systems rely heavily on the 
language priors. 
They are caused by the strong spurious correlation between certain question category and answers, e.g., the frequent co-occurrence of the question category  `what sport' and the answer `tennis' \citep{selvaraju2019taking}. As a result, the VQA models, which are over-reliant on the language priors of training set, fail to generalize to the OOD dataset, VQA-CP v2 \citep{agrawal2018don}.

\begin{figure}[t]
\centering
\includegraphics[width=0.47\textwidth]{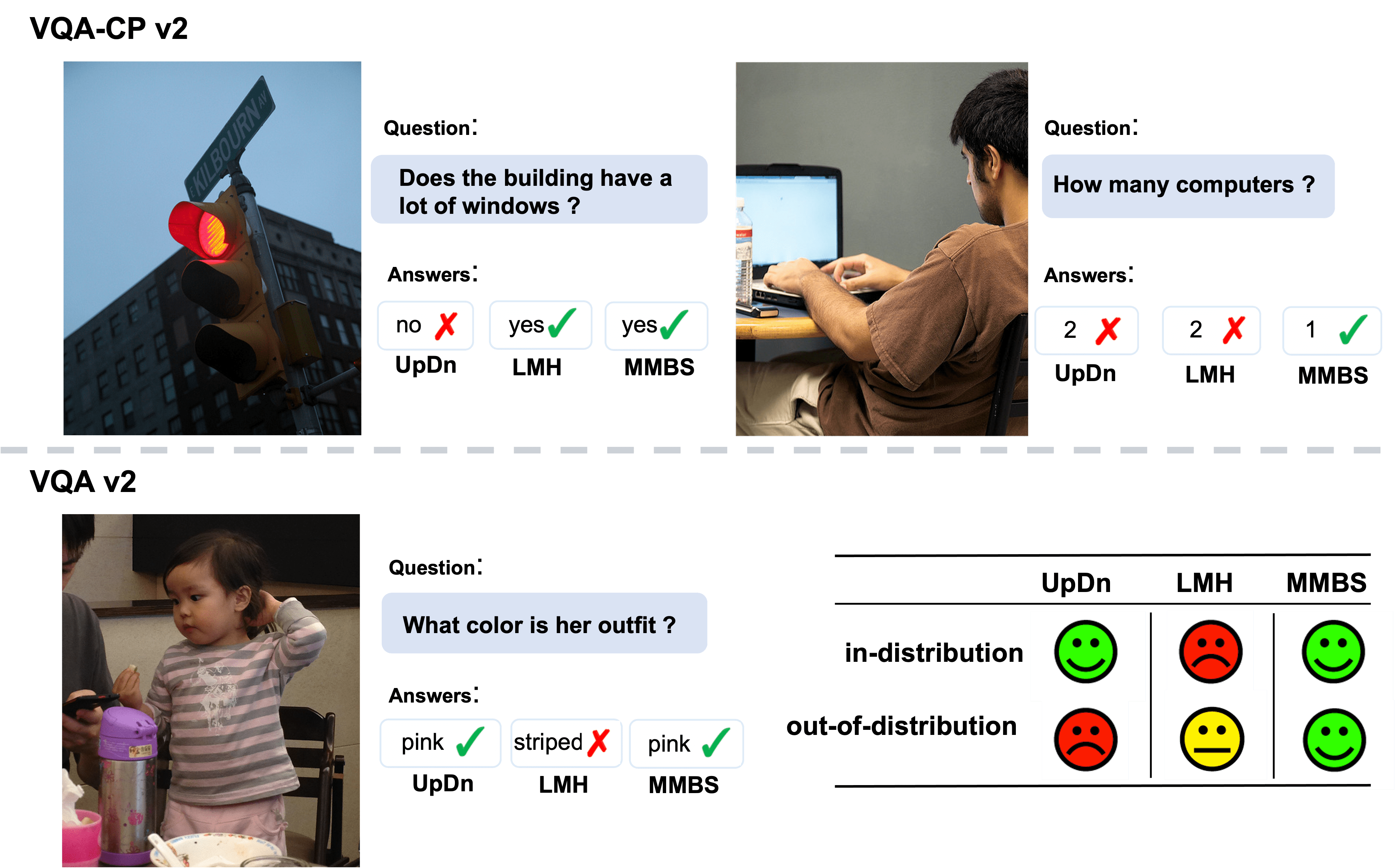} 
\caption{Qualitative comparison of our method LMH+MMBS against the plain method UpDn and the debiasing method LMH. In VQA-CP v2 (upper), the question types (`Does the' and `How many') bias UpDn to the most common answers (see Fig. \ref{ans_dis} for the answer distribution). LMH alleviates the language priors for yesno questions (upper left), while it fails on the more difficult non-yesno questions (upper right). Besides, LMH damages the ID performance, giving an uncommon answer to the common sample from VQA v2 (lower left). MMBS improves the OOD performance while maintains the ID performance (lower right). }
\label{qualitative}
\vspace{-0.3cm}
\end{figure}

Recently, several methods achieved remarkable progress in overcoming this language prior problem. 
They assign less importance to the biased samples that can be correctly classified with the spurious correlation. However, most of them achieve gains on VQA-CP v2 at the cost of degrading the model’s ID performance on the VQA v2 dataset (see Tab. \ref{ensem-sotas}). This trade-off suggests that the success of these methods merely comes from biasing the models to other directions, rather than endowing them with the reasoning capability and robustness to language priors. Ideally, a robust VQA system should maintain its performance on the ID dataset while overcoming the language priors, as shown in Fig. \ref{qualitative}.



We think the essence of both language-prior and trade-off problems is about the learning of biased samples. The former is caused by over-reliance on biased information from biased samples, while the latter is caused by undermining the importance of biased samples. Therefore, if a model can precisely exploit the biased samples for intrinsic information of the given task, both problems can be alleviated simultaneously.

Motivated by this, we propose a self-supervised contrastive learning method (MMBS) for building robust VQA systems by \textbf{M}ake the \textbf{M}ost of \textbf{B}iased \textbf{S}amples. 
Firstly, in view of the characteristics of the spurious correlations, we construct two kinds of positive samples for the questions of training samples to exploit the unbiased information, and then design four strategies to use the constructed positive samples. Next, we propose a novel algorithm to distinguish between biased and unbiased samples, so as to treat them differently. On this basis, we introduce an auxiliary contrastive training objective, which helps the model learn a more general representation with ameliorated language priors by narrowing the distance between original samples and positive samples in the cross-modality joint embedding space.


To summarize, our contributions are as follow:
\textbf{i}) We propose a novel contrastive learning method, which effectively addresses the language prior problem and the ID-OOD performance trade-off in VQA, by making the most of biased samples. \textbf{ii}) We propose an algorithm to distinguish between biased and unbiased samples and treat them differently in contrastive learning. \textbf{iii}) Experimental results demonstrate that our method is compatible with various VQA backbones and achieve competitive performance on the language-bias sensitive VQA-CP v2 dataset while preserving the original accuracy on the in-distribution VQA v2 dataset.


\section{Related Work}

\paragraph{Overcoming Language Priors in VQA.} \label{debiasingmethods}
Recently, the language biases in VQA datasets raised the attention of many researchers \citep{goyal2017making,antol2015vqa,agrawal2016analyzing,kervadec2021roses}. 
In response to this problem, numerous methods are proposed to debias the VQA models. The most effective ones of them can be roughly divided into two categories:
\textbf{Ensemble-based methods} \citep{ grand2019adversarial,belinkov2019don,cadene2019rubi,clark2019don,mahabadi2019simple,niu2021counterfactual}
introduce a biased model, which is designed to focus on the spurious features, to assist the training of the main model. For example, the recent method LPF \citep{liang2021lpf} leverages the output distribution of the bias model to down-weight the biased sample when computing the VQA loss. However, these methods neglect the useful information that helps reasoning in biased samples. 
\textbf{Data-balancing methods} \citep{zhu2020overcoming,liang2020learning} balance the training priors. 
For example, CSS and Mutant \citep{chen2020counterfactual,gokhale2020mutant}  generate samples by masking the critical object in images and word in questions and by semantic image mutations respectively. 
These methods usually outperform other debiasing methods with a large margin on VQA-CP v2, because they bypass the challenge of the imbalanced settings \citep{liang2021lpf,niu2021counterfactual} by explicitly balancing the answers' distribution at the training stage. 
Though our method constructs the positive questions, it does not change the training answers' distribution. We also extend our method to the data-balancing method SAR \citep{si2021check}. 

\paragraph{Contrastive Learning in VQA.}

Recently, the contrastive learning is well-developed in unsupervised learning \citep{oord2018representation,he2020momentum} while its application in VQA is still in initial stage. CL \citep{liang2020learning} is the first work to employ contrastive learning to improve VQA model’s robustness. Its motivation is to learn a better relationship among the input sample and the factual and counterfactual sample which are generated by CSS.  
However, CL brings weak OOD performance gain and ID performance drop based on CSS. In contrast, our method attributes the key point of solving language bias to the positive-sample designs for excluding the spurious correlations. It is model-agnostic and can boost models' OOD performance significantly while retain the ID performance.

\begin{figure}[t]
\centering
\includegraphics[width=0.46\textwidth]{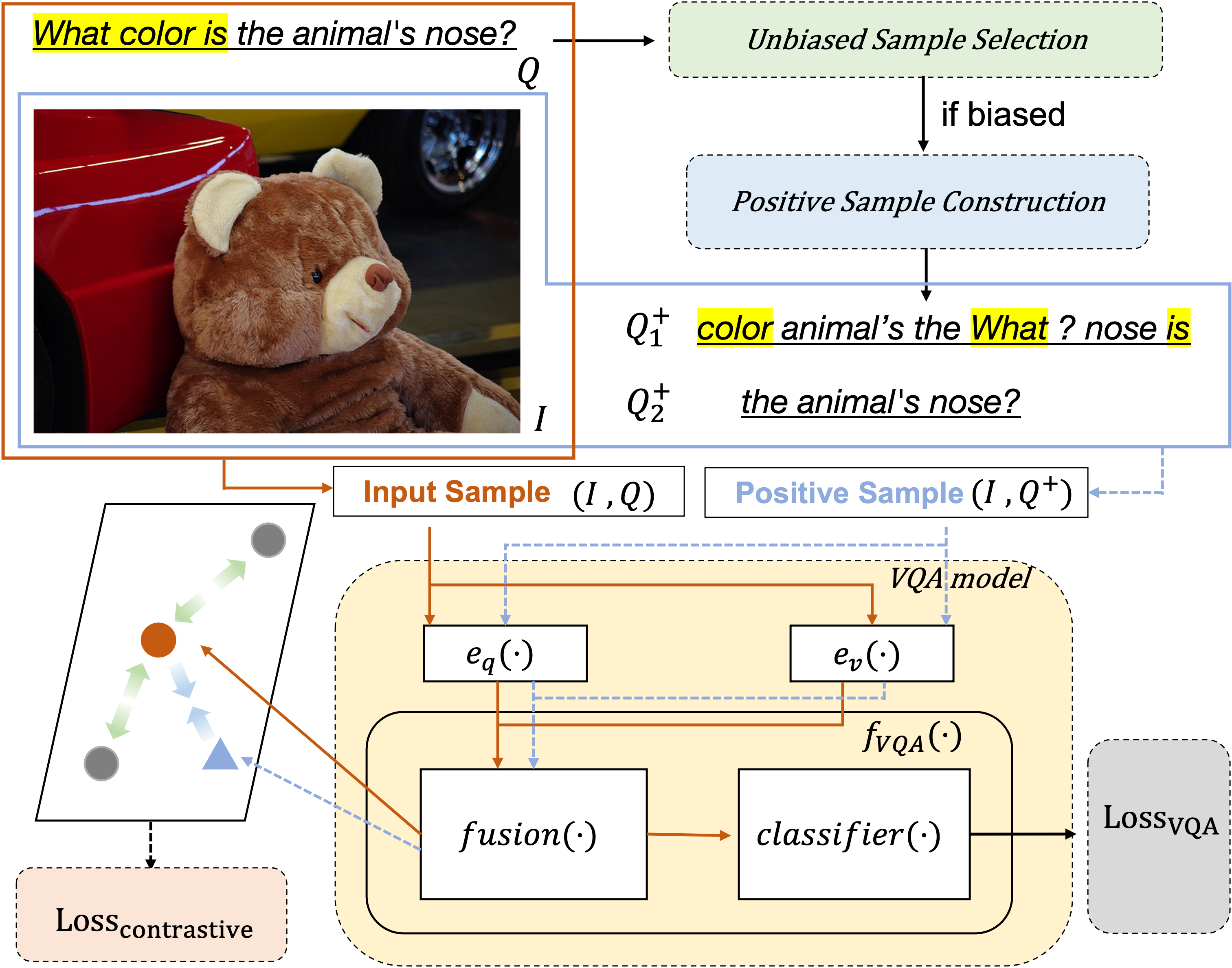} 
\caption{Overview of our method. The question category words are highlighted in yellow. The orange circle and blue triangle denote the cross-modality representations of the original sample and positive sample. The other samples in the same batch are the negative samples, which are denoted by the gray circles.}
\label{model}
\vspace{-0.2cm}
\end{figure}

\section{Method}
Fig. \ref{model} shows MMBS's overview, which includes: 1) A backbone VQA model; 2) A positive sample construction module; 3) An unbiased sample selection module; 4) A contrastive learning objective. 

\subsection{Backbone VQA Model} 
The backbone VQA model is a free choice in MMBS. The widely-used backbone models \citep{anderson2018bottom,mahabadi2019simple} treat VQA as a multi-class multi-label classification task. Concretely, given a VQA dataset $D=\{I_i,Q_i,A_i\}{_{i=1}^N}$ with $N$ samples, where $I_i \in I$, $Q_i \in Q$ are the image and question of the $i_{th}$ sample. $A_i \in A$ is the ground-truth answer which is usually in multi-label form, and  $tgt_i$ is the corresponding target score of each 
label. 
Most existing VQA models consist of four parts: the question encoder $e_q(\cdot)$, the image encoder $e_v(\cdot)$, the fusion function $F(\cdot)$ and the classifier $clf(\cdot)$. 
For example, LXMERT \citep{tan2019lxmert} encodes image and caption text separately to extract visual features $V_i = e_v (I_i) $, and textual features $T_i = e_q (Q_i) $ , in two streams. Next, the higher co-attentional transformer layers fuse the two features and project them into the cross-modality joint embedding space, i.e., $F(V_i, T_i)$. Finally, the classifier outputs the answer prediction: 
\begin{equation}
P(A|I_i, Q_i) = clf(F(V_i, T_i))
\end{equation}
The training objective minimizes the multi-label soft loss, $L_{vqa}$ , which can be formalized as follow:
\begin{equation}
\begin{aligned}
L_{vqa} = &-\frac{1}{N}\sum\nolimits_{i=1}^N[tgt_i \cdot log(\delta (F(V_i,T_i)))\\&+(1-tgt_i)\cdot log(1-\delta(F(V_i,T_i)))]
\end{aligned}
\end{equation}
where $\delta$ denotes the sigmoid function.

\subsection{Positive Sample Construction}\label{sr}
To make the most of the unbiased information contained in the biased sample, we first construct the positive samples which exclude the biased information. 
According to the construction of VQA-CP v2, there is a shift between the training and test set in terms of answer distribution under the same question category \citep{teney2020value,agrawal2018don}. As a result, the frequency co-occurrence of certain answer and question category in the training set produces a major source of bias. Therefore, we construct two kinds of positive questions ($Q_i^+$) by corrupting the question category information of each input question ($Q_i$):


\emph{\textbf{Shuffling}}: We randomly shuffle the words in the question sentence so that the question category words are mixed with the other words. This increases the difficulty of building the correlations between question category and answer.

\emph{\textbf{Removal}}: We remove the question category words from the question sentence. It eliminates the co-occurrence of answer and question category words completely. 

We notice that the construction process could induce some unexpected noise in the positive samples. To tackle this concern, we present more positive samples in \textbf{Appendix} \ref{app1.1} and discuss their quality and potential impact on our method.

We also propose four strategies for using the constructed positive questions during training:


\emph{\textbf{S}}: \emph{Use the \textbf{Shuffling} positive questions.}

\emph{\textbf{R}}:	\emph{Use the \textbf{Removal} positive questions.}

\emph{\textbf{B}}:	\emph{Use \textbf{both} positive questions.}

\emph{\textbf{SR}}:	\emph{Use the \textbf{Shuffling} positive questions for non-yesno (i.e., `Num' and `Other') questions and use the \textbf{Removal} ones for yesno (i.e., `Y/N') questions. }

The \emph{\textbf{SR}} strategy deals with yesno and non-yesno questions in different ways based on their characteristics. Intuitively, the question categories of the yesno questions usually contain little information, as they are mostly comprised of `is', `do', etc. By contrast, the question categories of non-yesno questions tend to contain more information which is important for answering correctly. Therefore, \emph{\textbf{Removal}} is not applied to non-yesno questions.

Adopting any strategy above, we can obtain the positive samples $\{I_i,Q_i^+\}{_{i=1}^B}$ for input samples$\{I_i,Q_i\}{_{i=1}^B}$. The negative samples $\{I_b,Q_b\}{_{b=1}^B}$, where $b\neq i$,  are the other samples in the same batch. $B$ is the batch size of training.

\subsection{Unbiased Sample Selection}
Following \citet{kervadec2021roses}, we define unbiased (or OOD) samples as the infrequent samples in the answers' distribution of each question category in training set. Therefore, the unbiased samples are unlikely to contain spurious correlations, which makes them beneficial to OOD robustness. Moreover, some unexpected noise in the positive samples may negatively impact the learning of unbiased samples.
For the above reasons, we do not construct positive samples for the unbiased samples. To filter out the unbiased samples, we propose a novel algorithm,
consisting of three steps: (i) calculating the answer frequencies; (ii) determining the unbiased answer proportion; (iii) selecting the unbiased samples.


\paragraph{Answer frequencies.}
We denote the $i_{th}$ sample's question category, ground truth answer and soft target score as $C_i \in C$ (65 categories in total), $A_i$ and $tgt_i$ 
respectively. We measure how frequent the answer $A_j$ appears in the question category $C_k$ as follows:
\begin{equation}
\begin{aligned}
    Freq_{C_k}^{A_j}& =  \sum\nolimits_{i=1}^{M_{C_k}}(tgt_i) \ ,\ if \ A_i=A_j
\end{aligned}
\label{eq3}
\end{equation}
where $M_{C_k}$ is the number of all samples with the same category $C_k$. If a sample has a multi-label answer $A_i$, we count each answer’s score respectively. A lower value of $Freq_{C_k}^{A_j}$ indicates weaker spurious correlations between $A_j$ and $C_k$, and thus the corresponding samples are deemed as unbiased. We introduce a hyper-parameter $\beta \in \begin{bmatrix} 0,1\end{bmatrix} $ to control the proportion of the unbiased samples.

\begin{figure}[t]
\centering

\includegraphics[width=0.46\textwidth]{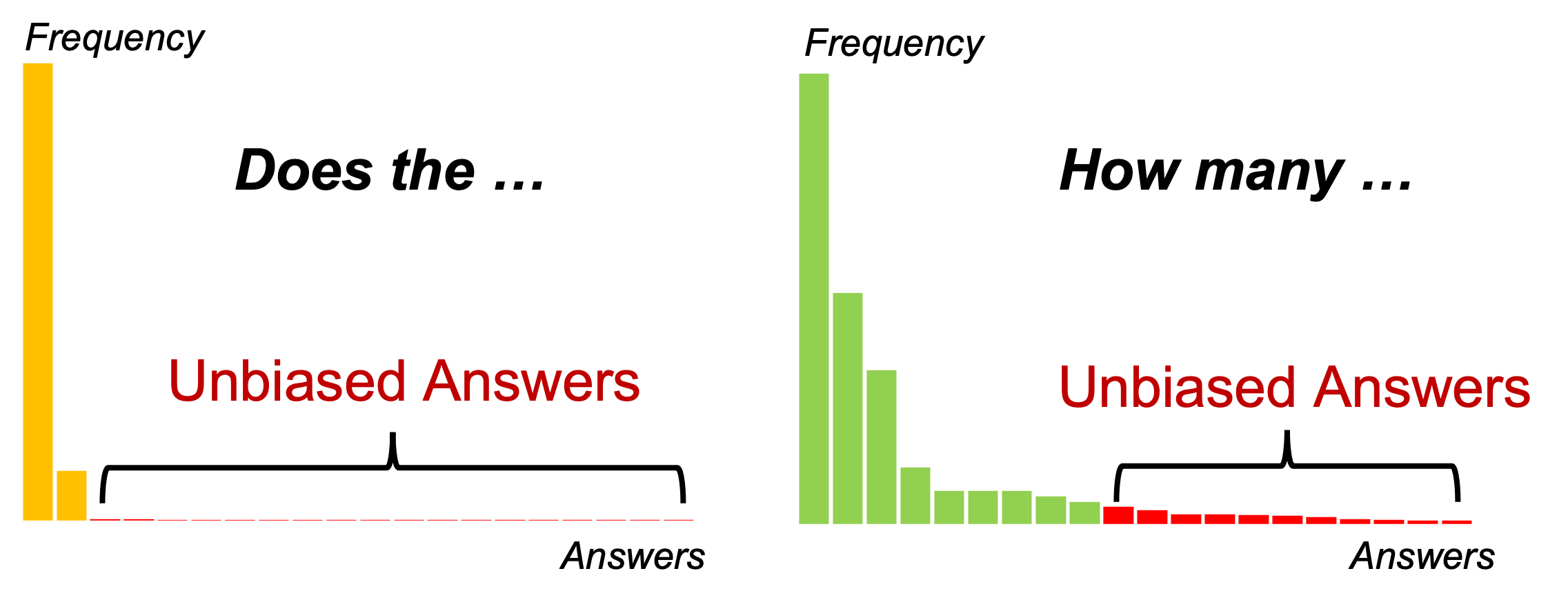} 
\vspace{-0.2cm}
\caption{The answers’ distributions of the yesno questions with “Does the” (left) and non-yesno questions with “How many” (right). The former has a low entropy and the latter has a high entropy.}
\vspace{-0.2cm}
\label{entropy}
\end{figure}


\paragraph{Entropy-based correction factor.} The answers' distributions of $|C|$ question categories are different. Empirically, when the entropy of an answers' distribution is lower, more answers will be associated with only a few samples, so that the unbiased answer proportion should be higher. Otherwise, it should be lower. An illustration is given in Fig. \ref{entropy}. Therefore, we propose an entropy-based correction factor $W_{C_k}$ to dynamically adjust the $\beta$ for each category $C_k$:

\begin{equation}
\begin{aligned}
&W_{C_k} = 1-sigmoid (E_{C_k}-mean(E)) 
\\
&E_{C_k} = Entropy (Freq_{C_k}/SUM)
\end{aligned}
\end{equation}
where  $E$ represents $\{E_{C_k}\}_{k=1}^{|C|}$ and $SUM$ represents the sum of $Freq_{C_k}$. 
When the entropy is lower, the $W_{C_k}$ is closer to 1, and otherwise $W_{C_k}$ is closer to 0.
Finally, we obtain the unbiased answer proportion $P_{C_k} = W_{C_k} * \beta $.

\paragraph{Selecting unbiased samples.}
For each question category $C_k$, we obtain a list of unbiased answers which rank in the last $P_{C_k}$ in $Freq_{C_k}$. Then we determine the samples whose ground truth (highest-score) answer belongs to this list as unbiased samples. The unbiased sample statistics are shown in \textbf{Appendix} \ref{app1.2}. 
If a sample is biased, we adopt the strategy mentioned in previous section to construct its positive sample. If it is unbiased, we use the original sample as its positive sample. 


\begin{table*}\footnotesize
    \centering
    \resizebox{0.95\linewidth}{!}{
    \begin{tabular}{cc|l|lllll|lllll}
    \hline 
    &&     &  \multicolumn{5}{c}{ VQA-CP v2 test }&  \multicolumn{5}{|c}{VQA v2 val }  \\ \cline{1-7} \cline{7-13}
    &&    Methods & All & Y/N & Num & Other & Gap $\uparrow$ & All & Y/N & Num & Other & Gap $\uparrow$ \\ \cline{3-13}
    \multicolumn{2}{c|}{\multirow{6}{*}{\rotatebox{90}{Plain Models}}} &    BAN  & 37.03 & 41.55 & 12.43 & 41.4 & \multirow{2}{*}{\textbf{+10.60}}  & 63.9 & 81.42 & 45.18 & 55.54 &  \multirow{2}{*}{\textbf{+0.88}}  \\ 
    &&    \textbf{\ \ \ \ \ +MMBS} & \textbf{47.63}&	\textbf{66.18}&	\textbf{16.36}	&\textbf{46.49}&		&\textbf{64.78}&	\textbf{82.03}&	\textbf{46.48}&	\textbf{56.51}&	 \\\cline{3-13}
    &&    UpDn    & 39.74 & 42.27 & 11.93 & 46.05 &\multirow{2}{*}{\textbf{+8.45}}  & 63.48 & \textbf{81.18} & 42.14 & 55.66 & \multirow{2}{*}{\textbf{+0.36}} \\
    && \textbf{\ \ \ \ \ +MMBS} &\textbf{48.19} & \textbf{65.00} & \textbf{14.05} & \textbf{48.75} &   & \textbf{63.84} & 79.61 & \textbf{44.23} & \textbf{57.05} & \\ \cline{3-13}
       
    &&    LXM  & 47.19&	50.55	&24.06&	51.77&	\multirow{2}{*}{\textbf{+9.32}}&\textbf{71.01}&	88.24&	54.07 &	\textbf{62.39} & \multirow{2}{*}{-0.16}\\  
    &&    \textbf{\ \ \ \ \ +MMBS} & \textbf{56.51} & \textbf{79.83} & \textbf{28.70} & \textbf{51.92} & &
       70.85 & \textbf{88.25} & \textbf{55.67} & 61.63 & \\ \hline
    
    \multicolumn{2}{c|}{ \multirow{3}{*}{\rotatebox{90}{Debiasing} \rotatebox{90}{Models}}} &    LMH  &  52.01 & 72.58 & 31.12 & 46.97 & \multirow{2}{*}{\textbf{+4.43 }} & 56.35 & 65.06 & 37.63 & 54.69 & \multirow{2}{*}{\textbf{+5.52 }} \\ 

    &&    \textbf{\ \ \ \ \ +MMBS} &\textbf{56.44} & \textbf{76.00} & \textbf{43.77} & \textbf{49.67} &  & \textbf{61.87} & \textbf{75.86} & \textbf{40.34} & \textbf{56.95} &  \\ \cline{3-13}

    &&    SAR& 66.73 & 86.00 & 62.34 & 57.84 &\multirow{2}{*}{\textbf{+1.66}}  & 69.22 & \textbf{87.46} & \textbf{51.20} & 60.12 & \multirow{2}{*}{\textbf{+0.21}}  \\ 
    &&    \textbf{\ \ \ \ \ +MMBS} &\textbf{68.39}&	\textbf{87.30}&	\textbf{65.21}&	\textbf{59.36}&	&\textbf{69.43}	&87.39	&50.37&	\textbf{60.82}&	\\ \hline
    \end{tabular}
    }
    \caption{Results on VQA-CP v2 test and VQA-v2 validation set based on different VQA models. `Gap' denotes the accuracy improvement of MMBS over the base model.} 
\label{backbones}
\end{table*}

\subsection{Contrastive Learning Objective}
Given input sample ($I_i ,Q_i$), we have the positive sample ($I_i ,Q_i^+$) and the negative samples { ($I_b ,Q_b$)}${_{b=1}^B}$ in the same batch, where $b\neq i$. After feeding them into the VQA model, we obtain the cross-modality fusion representation of the input sample, $F(V_i, T_i)$, positive sample $F(V_i, T_i^+)$ and negative samples { $F(V_b, T_b)$}$_{b=1}^B$,  which are denoted as the anchor $a$, the positive $p$ and the negative {$n_b$}$_{b=1}^B$ respectively. Following \citep{robinson2020contrastive,liang2020learning}, we use the cosine similarity, $cos(\cdot)$, as the scoring function. 
The contrastive loss \citep{oord2018representation} is formulated as:
\begin{equation}
   L_{cl} = \underset{a, p, n_b}{\mathbb{E}}\left[-\log
   \frac{e^{cos(a, p)}}{e^{cos(a, p)}+\sum_{b=1}^{B} e^{cos(a, n_b)}}\right]
\end{equation}

By minimizing it, the models can focus on the unbiased information from the positive question. The overall loss of MMBS is 
formulated as:
$L = L_{vqa} + \alpha * L_{cl}$
, where $\alpha$ is the weight of $L_{cl}$.

\subsection{Inference Process\label{inference}}

After training with this contrastive loss, the models can handle the question in \emph{\textbf{original}}, \emph{\textbf{Shuffling}} and \emph{\textbf{Removal}} forms (Sec. \ref{sr}) in the inference phase.\footnote{The models without MMBS performs much worse when the question is in  \emph{\textbf{Shuffling}} or \emph{\textbf{Removal}} forms.}
We find that in the framework of MMBS, \emph{\textbf{Shuffling}} can further boost OOD performance for the plain models (e.g., UpDn and LXM), while \emph{\textbf{original}} performs the best for debiasing methods (e.g., LMH, SAR). Therefore, we shuffle the question words at test time when applying MMBS to the plain models. Detailed discussions are shown in the next section. 



\section{Experiments}
\subsection{Datasets and Evaluation} 
We evaluate our models on the OOD VQA-CP v2  \citep{agrawal2018don} and the ID VQA v2 \citep{goyal2017making} with the standard evaluation metric \citep{antol2015vqa} based on accuracy. 
Previous works \citep{chen2020counterfactual,si2021check,gokhale2020mutant} think that a minor accuracy difference between VQA v2 and VQA-CP v2 shows the real robustness. 
This encourages the researchers to work in the direction that increases the accuracy on VQA-CP v2 by sacrificing the performance on VQA v2. However, a robust VQA model should perform well on both datasets. Therefore, we compute the relative accuracy between each method and its base method on both ID and OOD datasets.


\subsection{Baselines and Implementations}
Our approach is general to various VQA backbones. In the work, we evaluate MMBS based on three plain VQA models (which are not specially designed for overcoming language priors): \textbf{BAN}  \citep{kim2018bilinear}, \textbf{UpDn}  \citep{anderson2018bottom} and \textbf{LXMERT} (LXM), and two debiasing methods: \textbf{LMH} \citep{clark2019don} and \textbf{SAR} \citep{si2021check}. 

We also compare our methods with the state-of-the-art methods on VQA-CP v2, which contain: 1) The ensemble-based methods: \textbf{AdvReg.} \citep{ramakrishnan2018overcoming}, \textbf{GRL} \citep{grand2019adversarial},  \textbf{RUBi} \citep{cadene2019rubi}, \textbf{DLR} \citep{2020Overcoming}, \textbf{LMH} \citep{clark2019don}, \textbf{CF-VQA} \citep{niu2021counterfactual}, \textbf{LPF} \citep{liang2021lpf}. 2) The data-balancing methods: \textbf{SSL} \citep{zhu2020overcoming}, \textbf{CSS} \citep{chen2020counterfactual}, \textbf{CL} \citep{liang2020learning}, \textbf{SAR} \citep{si2021check} and \textbf{MUTANT} (best-performance method) \citep{gokhale2020mutant}. 

Following the baselines above, the checkpoint for evaluation is also picked by the test set directly in the work due to the lack of val set \citep{teney2020value,agrawal2018don}. In this paper, we mainly report the results with \textbf{\emph{SR}} strategy.  We also conduct experiments to analyze the impact of different positive-sample construction strategies. More implementation details are shown in \textbf{Appendix} \ref{app2}.
\subsection{Main Results}
\paragraph{Performance based on different VQA models.}
As can be seen in Tab. \ref{backbones}, regardless of the backbone architectures and debiasing methods, our proposed method consistently outperforms the baselines with comfortable margin (1.66 \textasciitilde 10.60 absolute accuracy improvement) on OOD VQA-CP v2. For the plain models, MMBS particularly improves the performance on yesno questions (22.73 \textasciitilde 29.28) because the simple yesno
questions are more susceptible to the influence of language bias \citep{zhu2020overcoming,liang2021lpf}. In terms of the ID dataset, the baselines' performance can also be also improved or at least maintained with MMBS, while most debiasing methods sacrifice the accuracy on VQA v2 (see the  corresponding column in Tab. \ref{ensem-sotas}). Especially, compared with LMH, LMH+MMBS gets a prominent accuracy boost of 5.52 on VQA v2. This is because making the most of biased samples can effectively alleviate the ID performance decline resulting from the debiasing method LMH.


\begin{table}\footnotesize
    \centering
    \resizebox{1.0\linewidth}{!}{
    

        
    \begin{tabular}{l|p{12pt}cl|p{12pt}r|l}
    \hline 
         &  \multicolumn{3}{c}{ VQA-CP v2 test }&  \multicolumn{2}{|c|}{VQA v2 val } &Gaps \\ \cline{1-4} \cline{4-6}
        Methods & All &\tiny \  Y/N  \  \ Num \ Other & Gap $\uparrow$ & All  & Gap $\uparrow$ &Sum \\ \hline
    
        UpDn    & 39.74 &\tiny\ 42.27 \ 11.93 \ 46.05 &  & 63.48 &  & \\
  \ +AdvReg.&	41.17 & \tiny\	65.49 \	15.48	\ 35.48&	+1.43&	62.75&	-0.73&+0.70 \\
  \ +GRL & 42.33&\tiny \  59.74  \   14.78 \  40.76 &+2.59 &51.92&-11.56&-9.00 \\
  \ +RUBi&	44.23&\tiny \	67.05 \	17.48 \ 39.61&	+4.49&	61.16&	-2.32	&+2.17\\
\ +DLR&	48.87 &\tiny\ 70.99 \ 18.72 \ 45.57&	+9.13&	57.96&	-5.52&	+3.61\\
    \ +LMH  & 52.01 & \tiny\ 72.58 \ 31.12 \ 46.97 & +12.27 & 56.35  & -7.13 &+5.14 \\ 
        \ +CF-VQA& 53.55&\tiny	\ 91.15	\ 13.03	\ 44.97&	+13.81&	63.54&	+0.06	&+13.87\\
\ +LPF&	55.34&	\tiny\ 88.61	\ 23.78 \ 46.57&	+15.60&	55.01&	-8.47&	+7.13 \\
    \cline{2-7}
        +LMH+\textbf{MMBS} & \textbf{56.44} &\tiny \ 76.00 \ 43.77 \ 49.67 & \textbf{+16.70} & 61.87  & -1.61 & \textbf{+15.09} \\  \hline

        LXM  & 47.19& \tiny\	50.55	\ 24.06 \ 51.77& &71.01& &	 \\  
\ +LMH* & 63.34 & \tiny\ 78.28 \ 65.95 \ 54.79 &+16.15 & 69.49 & -1.52&+14.63 \\
\    +U-SAR*	&64.98 &\tiny\ 81.89	\ 59.65	\ 57.61	&+17.79	 &69.17&-1.84&+15.95 \\   \cline{2-7}
+LMH+\textbf{MMBS}	& \textbf{65.70}&\tiny \ 81.70 \ 61.24 \ 58.54&	\textbf{+18.51} &	70.29 &	-0.72 &	\textbf{+17.79} \\ 
+U-SAR+\textbf{MMBS}&	\textbf{68.01}& \tiny\	86.55 \	64.69 	\ 59.21&	\textbf{+20.82}	 &69.29&-1.72& \textbf{+19.10}\\

        \hline
    \end{tabular}
    }
    \caption{Comparison with the state-of-the-art ensemble-based methods.  `Gap' denotes the accuracy improvement of the debiasing methods over their base models.  * denotes the strong baselines introduced in this paper.}
    \vspace{-0.2cm}
\label{ensem-sotas}
\end{table}

\vspace{-0.2cm}
\paragraph{Comparison with ensemble-based SOTAs.} The upper part of Tab. \ref{ensem-sotas} compares the methods based on the UpDn backbone. We can observe that: 1) Compared with UpDn, most ensemble-based methods suffer from obviously performance drops on VQA v2. This phenomenon attests to the trade-off between the ability to overcome the language priors and the ability to memorize the knowledge of in-distribution samples. Though to a certain extent, CF-VQA alleviates the phenomenon, its accuracy on VQA-CP v2 is prominently lower than our method. 2) LMH+MMBS performs the best on VQA-CP v2 and rivals the accuracy of the backbone on VQA v2, clearly surpassing the previous best in `GapsSum'. This shows that the trade-off problem is effectively alleviated by the propose method. 3) The previous methods, e.g., CF-VQA and LPF, achieve high accuracy on the simple yesno question where the language biases are more likely to exist. By contrast, our method substantially improves over them on the more challenging non-yesno question, while achieves relatively good performance on the yesno questions. 

The methods in the lower part of Tab. \ref{ensem-sotas} are based on the LXM backbone. LXM is a cross-modal pre-trained model that has been used as backbone in some data-balancing method to further boost performance \citep{si2021check,gokhale2020mutant}. However, the performance of LXM with ensemble-based methods has not been fully investigated. We introduce two strong baselines based on LXM, i.e., LXM+LMH and U-SAR. LXM+LMH represents the LXM model trained with LMH method, which is widely used as an essential component by existing methods \citep{chen2020counterfactual,liang2020learning,si2021check}. U-SAR is a variants of the two-stage method \textbf{SAR},
with the data-balancing method SSL replaced with UpDn. We can see that MMBS further promotes the two strong baselines, enhancing the OOD performance and relieving the ID performance drop. Moreover, the LXM-based MMBS is even competitive with the data-balancing methods that generate samples.


\begin{table}\footnotesize
    \centering
    \resizebox{0.95\linewidth}{!}{
    \begin{tabular}{l|l|cc|cc|c} \hline

         & & \multicolumn{2}{c}{ \small VQA-CP v2 test }&  \multicolumn{2}{|c|}{VQA v2 val } &Gaps  \\ 
       Methods &Base & All &Gap$\uparrow$ & All &Gap$\uparrow$ &Sum \\      \hline
    
SSL & UpDn &	57.59& +17.85&	63.73 &+0.25&+18.10\\
LMH+CCS	& UpDn&  58.95&	+19.21&59.91&-3.57&+15.64 \\ 
LMH+CCS+CL& UpDn&  	59.18& +19.44&	57.29&-6.19 &+13.25 \\ 
SAR&LXM&	66.73&+19.54&	69.22&-1.79 & +17.75 \\ 
MUTANT&	LXM& 69.52&+22.33&	70.24 &-0.77 &+21.56\\ \hline
SAR+\textbf{MMBS} &LXM&	 68.39&+21.20&69.43 &-1.58&+19.62\\  \hline
        
    \end{tabular}
    }
    \caption{Comparison with the state-of-the-art data-balancing methods.}
    \vspace{-0.2cm}
\label{databalancing}
\end{table}


\vspace{-0.1cm}
\paragraph{Comparison with data-balancing SOTAs.}

We can derive three observations from the results in Tab. \ref{databalancing}: 1) Most data-balancing methods also hurt the ID performance, which is the result of a mismatch between the balanced training priors and the biased test priors. 
2) Another existing contrastive learning model LMH+CSS+CL \citep{liang2020learning}, which can only be applied to the data-balancing method LMH+CSS, achieves a mild improvement of 0.23 on VQA-CP v2 and sacrifices the accuracy on VQA v2. Compared with it, our MMBS is general to various VQA backbones and does not hurt the ID performance. 3) Our SAR+MMBS brings encouraging performance gain over the strong baseline SAR and achieves competitive performance against the best-performing method MUTANT without utilizing extra manual annotations to construct extensive data.

\begin{table}\footnotesize
    \centering
    \resizebox{0.95\linewidth}{!}{
    \begin{tabular}{l|l|llll}
    \hline
        Method & Strategy & All & Y/N & Num & Other \\ \hline
      UpDn    & Base* & 41.06 & 43.13 & 13.71 & 47.48 \\ 
         & \emph{\textbf{S}} & 42.26 & 45.11 & 13.99 & 48.52 \\ 
         & \emph{\textbf{R}} & 42.83 & 57.74 & 12.25 & 43.41 \\ 
         & \emph{\textbf{B}} & 44.37 & 51.58 & \textbf{14.94} & 48.67 \\
         & \emph{\textbf{SR}} & \textbf{48.19} & \textbf{65.00} & 14.05 & \textbf{48.75} \\ \hline
       LXM  & Base* & 47.19 & 50.55 & 24.06 & 51.77 \\ 
         & \emph{\textbf{S}} & 47.90 & 52.71 & 26.48 & 51.26 \\ 
         & \emph{\textbf{R}} &  52.11 & 63.65 & 27.89 & \textbf{52.72} \\ 
         & \emph{\textbf{B}} & 50.76 & 61.33 & \textbf{29.21} & 51.14 \\ 
         & \emph{\textbf{SR}} & \textbf{56.51} & \textbf{79.83} & 28.70 & 51.92 \\ \hline
      LMH   & Base* & 52.58 & 67.10 & 36.59 & 49.36 \\ 
         & \emph{\textbf{S}} & 55.89 & 76.67 & 37.64 & 50.01 \\ 
         & \emph{\textbf{R}} & 55.87 & \textbf{76.79} & 34.96 & \textbf{50.65} \\
         & \emph{\textbf{B}} & 55.62 & 76.47 & 35.71 & 50.15 \\ 
         & \emph{\textbf{SR}} & \textbf{56.44} & 76.00 & \textbf{43.77} & 49.67 \\ \hline
         
    \end{tabular}
    }
    \caption{Results of different positive-sample construction strategies on the VQA-CP v2 test set.}
\vspace{-0.2cm}
\label{Strategy}
\end{table}

\subsection{Analysis on Individual Components and Hyper-Parameters}
\paragraph{The effect of positive sample construction strategies.}
As shown in Tab. \ref{Strategy}, we conduct experiments based on three widely used methods, i.e., the plain model UpDn, pre-trained model LXM and UpDn with the debiasing method LMH. From the results UpDn and LXM, we can observe that: 1) Both \emph{\textbf{S}} and \emph{\textbf{R}} strategies gain performance boost. This shows that the designs of both of them are sound and effective, and their benefits outweigh the potential semantic noise. 2) \emph{\textbf{R}} strategy has a better overall performance than \emph{\textbf{S}} because the model may still learn the superficial correlation between answer and the question category even when the category words are shuffled with the other words of the sentence. 
3) \emph{\textbf{SR}} strategy performs the best among the four strategies, especially on the yesno questions. The reason is that \emph{\textbf{R}} strategy significantly outperforms \emph{\textbf{S}} strategy on the yesno questions while the \emph{\textbf{S}} strategy performs well on the non-yesno questions. \emph{\textbf{SR}} strategy combines the advantages of both strategies. 4) \emph{\textbf{B}} strategy is obviously inferior to the \emph{\textbf{SR}} strategy. This is because learning from two positive samples for each sample simultaneously may confuse the model.  


From the results of LMH, we find that all the strategies considerably boost the performance, including the \emph{\textbf{S}} strategy. 
This is because the unbiased information contained in biased samples, which is useful for reasoning, is also being neglected by the ensemble-based methods. 
Through the contrastive learning objective, both \emph{\textbf{Shuffling}} and \emph{\textbf{Removal}} positive samples give them another channel to learn and utilize the useful information. \emph{\textbf{SR}} strategy still has the best performance among all the strategies. 


\begin{figure}[t]
\centering
\includegraphics[width=0.48\textwidth]{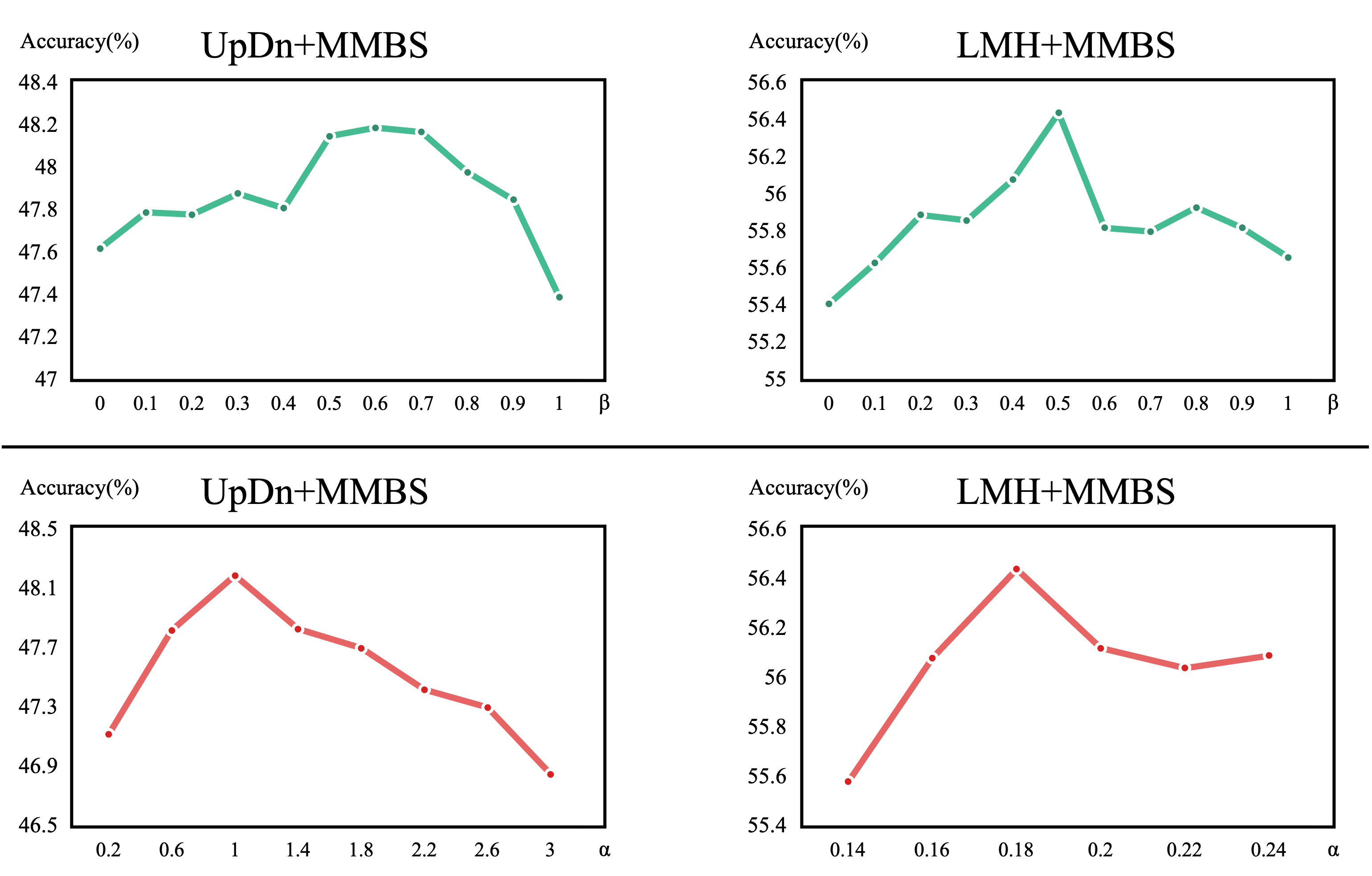} 
\caption{Results of UpDn+MMBS and LMH+MMBS on VQA-CP v2 with varying of $\beta$ (upper) and $\alpha$ (lower).} 
\label{beta_alpha}
\end{figure}

\paragraph{The effect of $\beta$ and $\alpha$.}
As shown in the upper plots of Fig. \ref{beta_alpha}, the accuracy rises first and then decreases as $\beta$ increases. There is a trade-off behind this phenomenon: when $\beta$ is too small, the method will construct the positive samples for the unbiased samples, which may affect the learning of robust information from the unbiased samples. When $\beta$ is too large, the method will not construct positive samples for some biased samples. This demeans the profits from the contrastive learning objective. 

The lower plots of Fig. \ref{beta_alpha} also revel a trade-off with the increase of $\alpha$. This suggests that the contrastive learning objective is beneficial but paying too much attention to this objective hurts the final performance. we also find that the best $\alpha$ for LMH+MMBS is smaller than that for UpDn+MMBS. This is because LMH itself already has certain ability to alleviate language priors.

\begin{table}\footnotesize
    \centering
    \resizebox{0.95\linewidth}{!}{
    \begin{tabular}{l|llll}
    \hline
        Method & All & Y/N & Num & Other \\ \hline
        UpDn & 41.06 & 43.13 & 13.71 & 47.48 \\
        UpDn\emph{+SR} & 47.62 & 62.72 & 13.92 & \textbf{48.95} \\
        UpDn\emph{+SR+$\beta$} & 48.00&	64.06	&14.10&	48.89   \\
        UpDn\emph{+SR+$\beta$+$W_C$}  & \textbf{48.19} & \textbf{65.00} & 14.05 & 48.75 \\ \hline

        LXM & 47.19 & 50.55 & 24.06 & 51.77 \\ 
        LXM\emph{+SR }& 55.26 & 77.13 & 27.33 & 51.47 \\ 
        LXM\emph{+SR+$\beta$ }&55.66&	78.64&	28.10&	51.17  \\
        LXM\emph{+SR+$\beta$+$W_C$ } & \textbf{56.51} & \textbf{79.83} & 28.70 & 51.92 \\ \hline
        LMH & 52.01 & 72.58 & 31.12 & 46.97 \\ 
        LMH\emph{+SR }& 55.41 & 76.50 & 37.20 & 49.35 \\ 
        LMH\emph{+SR+$\beta$ }&  56.15&	\textbf{77.46}&	37.90&	50.00  \\
        LMH\emph{+SR+$\beta$+$W_C$ } & \textbf{56.44} & 76.00 & \textbf{43.77} & 49.67 \\ \hline

    \end{tabular}
    }
\caption{Results of ablation study on VQA-CP v2.}
\label{AblationStudy}
\end{table}

\begin{table}\footnotesize
    \centering
    \resizebox{0.95\linewidth}{!}{
    \begin{tabular}{l|l|llll}
    
    \hline
        Method & Form & \emph{\textbf{S}} & \emph{\textbf{R}} & \emph{\textbf{B}} & \emph{\textbf{SR}} \\  \hline
        UpDn & \emph{\textbf{original}} & 42.20 & 42.38 & 42.69 & 42.80 \\ 
         & \emph{\textbf{Shuffling}} & \textbf{42.26} & 33.68 & \textbf{44.37} & \textbf{48.19} \\  
         & \emph{\textbf{Removal}} & 26.15 & \textbf{42.83} & 43.19 & 22.67 \\ \hline
        LMH & \emph{\textbf{original}} & \textbf{55.89} & \textbf{55.87} & \textbf{55.62} & \textbf{56.44} \\  
         & \emph{\textbf{Shuffling}} & 54.14 & 39.93 & 52.3 & 52.64 \\  
         & \emph{\textbf{Removal}} & 31.46 & 49.4 & 47.48 & 32.43 \\ \hline
    \end{tabular}
    }
\caption{Results of UpDn+MMBS and LMH+MMBS with three question forms at test on VQA-CP v2. \emph{\textbf{S}}, \emph{\textbf{R}}, \emph{\textbf{B}} and \emph{\textbf{SR}} are the four strategies to use positive sample in training.}
\label{QForm}
\end{table}

\paragraph{Ablation study.}
Tab. \ref{AblationStudy} investigates the effect of each component of MMBS, i.e., the backbone models, the positive-sample construction module (\emph{\textbf{SR}}) and the unbiased sample selection module ($\beta$) which includes the correction factor $W_C$.
We find that: 1) \emph{+SR} constantly outperforms the base models significantly, especially on the yesno questions where the language biases tend to exist. We also conduct experiments for further validation of the effectiveness of the \emph{\textbf{SR}} strategy in \textbf{Appendix} \ref{app3}. 
2) Comparing the performance of \emph{+SR} and \emph{+SR+$\beta$}, we can find that the unbiased sample selection module always benefits MMBS. This attests to the intuition that we do not need to construct the positive samples for the unbiased samples. 3) The correction factor $W_C$ consistently has a positive impact on the model performance. This further demonstrates that dynamically adjusting the unbiased sample proportion for each question category is a useful strategy.


\subsection{Performance with different question forms at test.}\label{q_form_sec}

After contrastive learning using the positive questions, the models trained with MMBS can also take the positive question as input in the inference phase, while normal models cannot. For more comprehensive analysis, we report the results of three question forms here. Because the annotation of question categories should not be available at test, the \emph{\textbf{Removal}} questions are not used in the other experiments. From the results shown in Tab. \ref{QForm}, we find that: 1) For UpDn with the \emph{\textbf{S}}, \emph{\textbf{B}} and \emph{\textbf{SR}} strategies (which involve the \emph{\textbf{Shuffling}} positive sample), the performance is the best when the test question is in the \emph{\textbf{Shuffling}} form. This shows that the \emph{\textbf{Shuffling}} form input question, when used in the test stage, may further prevent the model from relying on the superficial correlations. 2) For LMH, when the input question during test is \emph{\textbf{original}}, the models always perform the best. This is probably because the LMH+MMBS method is robust enough and will not be easily biased by the superficial correlations in the \emph{\textbf{original}} questions. On the in-distribution settings, all the models obtain the best performance on VQA v2 when the test questions are in the \emph{\textbf{original}} form.

\subsection{Qualitative Analysis on the Effectiveness of MBSS}
\paragraph{Visualization of the answers' distribution.}

\begin{figure}[t]
\centering
\includegraphics[width=0.48\textwidth]{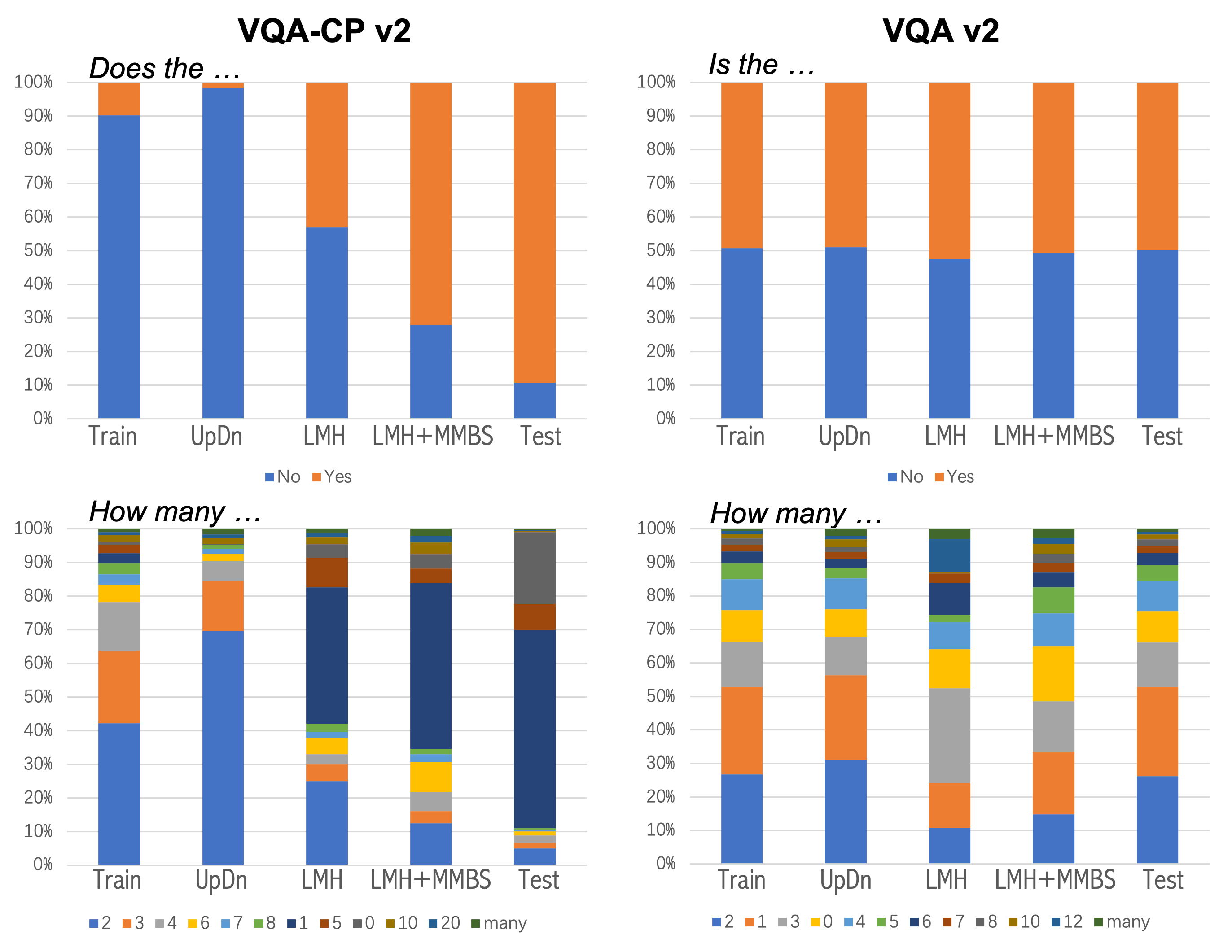} 
\caption{The answer distribution of the training sets, test sets, and three methods.}
\label{ans_dis}
\end{figure}
To better understand the effectiveness of MBSS, we compare the distribution of the predicted answers by three methods, i.e., UpDn, LMH and LMH+MMBS, and the real answer distribution of the training and test sets of VQA-CP v2 (left) and VQA v2 (right) in Fig. \ref{ans_dis}. From the left part, we find that UpDn tends to output the most frequent answers of training set, which demonstrates that it overfits the training priors. In comparison, LMH alleviates the domination of the biased answers and MBSS further mitigates the impact training priors, resulting in answer distributions that are closet to the test set. This explains why MBSS generalizes the best to the OOD VQA-CP v2 test set.

From the upper right plot, we see that for the relatively easy yesno question `Is the', when the training set is balanced in answer distribution, the three methods can also produce balanced answer distributions similar to the test set. For the question type `How many' on VQA v2, the most frequent answers in the training set, i.e., `2' and `1', account for much smaller proportion in the answer distribution of LMH. This is because that LMH diminishes the training signal from biased samples. Consequently, LMH performs worse on VQA v2 where most questions can be correctly answered by the common answers. By contrast, our method exploits the biased samples using contrastive learning rather than undermining them like LMH, and thus MBSS recovers the answers' distribution of ID test set.



\begin{figure}[t]
\centering
\includegraphics[width=0.48\textwidth]{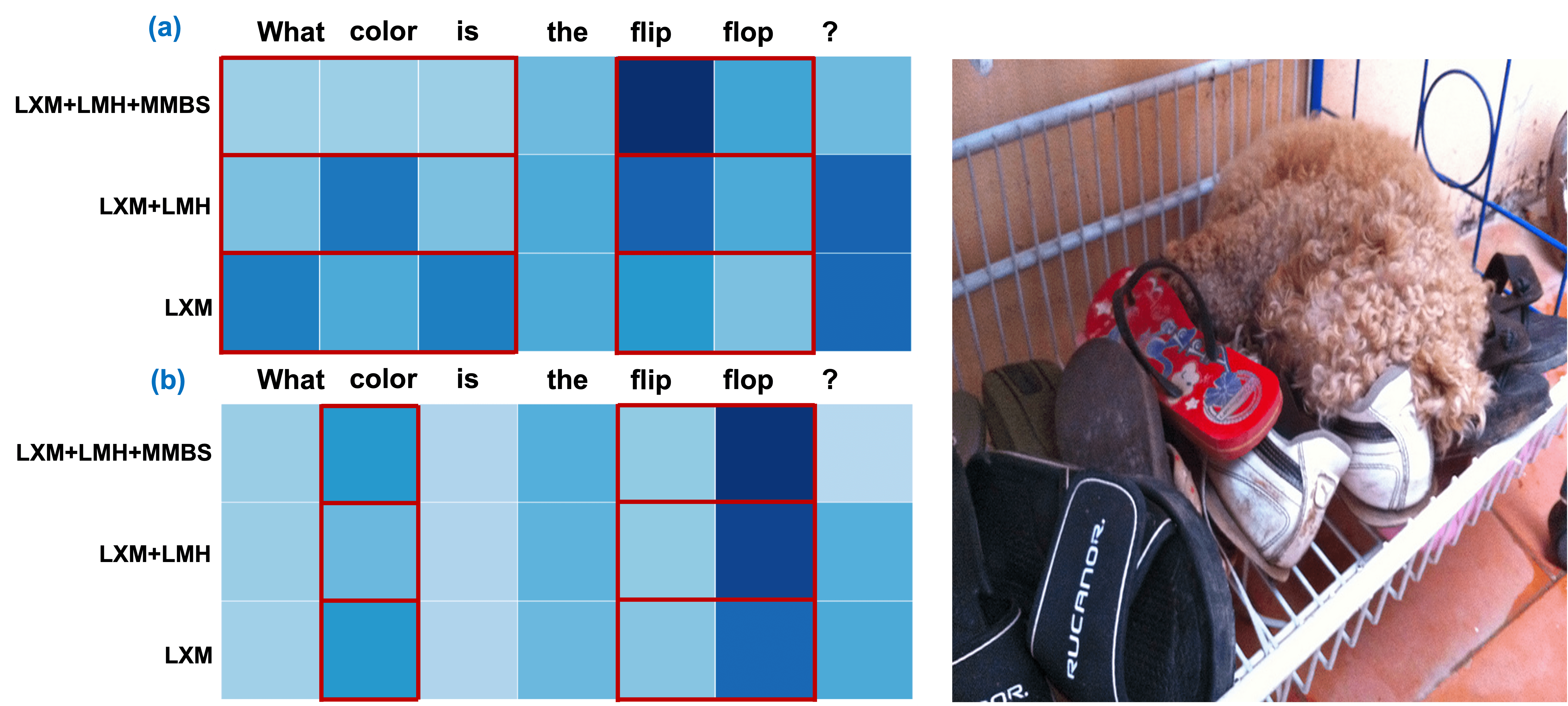} 
\caption{(a) The attention graph of the last cross-attention of cross-modality encoder, which averages the attention of all visual regions to each question word. (b) The attention graph of the last self-attention layer of the language encoder.}
\label{heatmap}
\end{figure}
\paragraph{Attention graph of question words.}
The attention graphs of LXM+LMH+MMBS, LXM+LMH and LXM are shown in Fig \ref{heatmap}. As highlighted in the red boxes, we focus on the question category words, i.e., `What color is' or `color', and the subject words, i.e., `flip flop'. We observe that: 1) For the cross-modality encoder (a) that extracts higher level representation for classification, LXM pays low attention to the subject words and high attention to the question category words, which is the source of language bias. In comparison, the introduction of LMH alleviates this problem and MBSS further shifts the attention to the subject words, which contain less biased information and have more specific visual groundings. 2) For the question encoder (b) that summarizes information from the textual domain, LXM+LMH pays less attention to the question category word `color', as compared with the other two methods. We conjecture that this can partly explain the poor performance of LMH on the ID dataset that contains strong language priors, because the word `color' is essential to the meaning of the question. LXM pays more attention to `color' but relatively less attention to the subject words. By contrast, our method assigns sufficient attention to both the question category and subject words, which can produces a better question representation.

\section{Conclusion}
In this paper, we propose a novel contrastive learning method to ameliorate the ID-OOD trade-off problem faced by most existing debaising methods for VQA models. Instead of undermining the importance of the biased samples, our method makes the most of them via contrastive learning. Considering the characteristics of language priors, we design the positive samples which eliminate the biased information. On this basis, we investigate several strategies to use the positive samples and design an algorithm that treat biased and unbiased samples differently in contrastive learning. The proposal is compatible with multiple backbone models and debiasing methods, and achieves competitive performance on OOD VQA-CP v2 while maintaining the performance on ID VQA v2. Meanwhile, our approach provides insights on how to avert the trade-off between in-distribution and out-of-distribution performance.


\section{Limitations}
\citeauthor{teney2020value} point out some practical issues in the use of VQA-CP v2, which has become the current OOD benchmark in VQA. These issues widely exist in the most of recent works (e.g., RUBi\citep{cadene2019rubi}, LMH\citep{clark2019don}, GRL\citep{grand2019adversarial}, DLR\citep{2020Overcoming}, AdvReg.\citep{ramakrishnan2018overcoming}, SAR\citep{si2021check}, SCR\citep{wu2019self}, MUTANT\citep{gokhale2020mutant}, etc.). Our method also suffers from them. Specifically, 1) our method is designed for the known biases (i.e., language priors) and the known construction of OOD splits of VQA-CP v2 (i.e., the inverse distribution shifts under the same question category between test and training sets).  Therefore, once the bias is unknown, or the training and test sets do not conform to such a construction procedure, MMBS may fail to generalize. 
2) Following all the baselines listed in \textbf{Sec. 4.2}, the checkpoint for evaluation is also picked by the test set directly in the work due to the lack of the val set of VQA-CP v2. Admittedly, an OOD benchmark with a val set is needed to standardize the OOD testing for VQA community.

\section*{Acknowledgments}
This work was supported by National Natural Science Foundation of China (No. 61976207, No. 61906187)

\bibliography{anthology,custom}

\begin{thebibliography}{31}
\expandafter\ifx\csname natexlab\endcsname\relax\def\natexlab#1{#1}\fi

\bibitem[{Agrawal et~al.(2016)Agrawal, Batra, and
  Parikh}]{agrawal2016analyzing}
Aishwarya Agrawal, Dhruv Batra, and Devi Parikh. 2016.
\newblock Analyzing the behavior of visual question answering models.
\newblock \emph{arXiv preprint arXiv:1606.07356}.

\bibitem[{Agrawal et~al.(2018)Agrawal, Batra, Parikh, and
  Kembhavi}]{agrawal2018don}
Aishwarya Agrawal, Dhruv Batra, Devi Parikh, and Aniruddha Kembhavi. 2018.
\newblock Don't just assume; look and answer: Overcoming priors for visual
  question answering.
\newblock In \emph{Proceedings of the IEEE Conference on Computer Vision and
  Pattern Recognition}, pages 4971--4980.

\bibitem[{Anderson et~al.(2018)Anderson, He, Buehler, Teney, Johnson, Gould,
  and Zhang}]{anderson2018bottom}
Peter Anderson, Xiaodong He, Chris Buehler, Damien Teney, Mark Johnson, Stephen
  Gould, and Lei Zhang. 2018.
\newblock Bottom-up and top-down attention for image captioning and visual
  question answering.
\newblock In \emph{Proceedings of the IEEE conference on computer vision and
  pattern recognition}, pages 6077--6086.

\bibitem[{Antol et~al.(2015)Antol, Agrawal, Lu, Mitchell, Batra, Zitnick, and
  Parikh}]{antol2015vqa}
Stanislaw Antol, Aishwarya Agrawal, Jiasen Lu, Margaret Mitchell, Dhruv Batra,
  C~Lawrence Zitnick, and Devi Parikh. 2015.
\newblock Vqa: Visual question answering.
\newblock In \emph{Proceedings of the IEEE international conference on computer
  vision}, pages 2425--2433.

\bibitem[{Belinkov et~al.(2019)Belinkov, Poliak, Shieber, Van~Durme, and
  Rush}]{belinkov2019don}
Yonatan Belinkov, Adam Poliak, Stuart~M Shieber, Benjamin Van~Durme, and
  Alexander~M Rush. 2019.
\newblock Don't take the premise for granted: Mitigating artifacts in natural
  language inference.
\newblock \emph{arXiv preprint arXiv:1907.04380}.

\bibitem[{Cadene et~al.(2019)Cadene, Dancette, Cord, Parikh
  et~al.}]{cadene2019rubi}
Remi Cadene, Corentin Dancette, Matthieu Cord, Devi Parikh, et~al. 2019.
\newblock Rubi: Reducing unimodal biases for visual question answering.
\newblock \emph{Advances in neural information processing systems},
  32:841--852.

\bibitem[{Chen et~al.(2020)Chen, Yan, Xiao, Zhang, Pu, and
  Zhuang}]{chen2020counterfactual}
Long Chen, Xin Yan, Jun Xiao, Hanwang Zhang, Shiliang Pu, and Yueting Zhuang.
  2020.
\newblock Counterfactual samples synthesizing for robust visual question
  answering.
\newblock In \emph{Proceedings of the IEEE/CVF Conference on Computer Vision
  and Pattern Recognition}, pages 10800--10809.

\bibitem[{Clark et~al.(2019)Clark, Yatskar, and Zettlemoyer}]{clark2019don}
Christopher Clark, Mark Yatskar, and Luke Zettlemoyer. 2019.
\newblock Don't take the easy way out: Ensemble based methods for avoiding
  known dataset biases.
\newblock \emph{arXiv preprint arXiv:1909.03683}.

\bibitem[{Gokhale et~al.(2020)Gokhale, Banerjee, Baral, and
  Yang}]{gokhale2020mutant}
Tejas Gokhale, Pratyay Banerjee, Chitta Baral, and Yezhou Yang. 2020.
\newblock Mutant: A training paradigm for out-of-distribution generalization in
  visual question answering.
\newblock \emph{arXiv preprint arXiv:2009.08566}.

\bibitem[{Goyal et~al.(2017)Goyal, Khot, Summers-Stay, Batra, and
  Parikh}]{goyal2017making}
Yash Goyal, Tejas Khot, Douglas Summers-Stay, Dhruv Batra, and Devi Parikh.
  2017.
\newblock Making the v in vqa matter: Elevating the role of image understanding
  in visual question answering.
\newblock In \emph{Proceedings of the IEEE Conference on Computer Vision and
  Pattern Recognition}, pages 6904--6913.

\bibitem[{Grand and Belinkov(2019)}]{grand2019adversarial}
Gabriel Grand and Yonatan Belinkov. 2019.
\newblock Adversarial regularization for visual question answering: Strengths,
  shortcomings, and side effects.
\newblock \emph{NAACL HLT 2019}, page~1.

\bibitem[{He et~al.(2020)He, Fan, Wu, Xie, and Girshick}]{he2020momentum}
Kaiming He, Haoqi Fan, Yuxin Wu, Saining Xie, and Ross Girshick. 2020.
\newblock Momentum contrast for unsupervised visual representation learning.
\newblock In \emph{Proceedings of the IEEE/CVF Conference on Computer Vision
  and Pattern Recognition}, pages 9729--9738.

\bibitem[{Jing et~al.(2020)Jing, Wu, Zhang, Jia, and Wu}]{2020Overcoming}
C.~Jing, Y.~Wu, X.~Zhang, Y.~Jia, and Q.~Wu. 2020.
\newblock Overcoming language priors in vqa via decomposed linguistic
  representations.
\newblock \emph{Proceedings of the AAAI Conference on Artificial Intelligence},
  34(7):11181--11188.

\bibitem[{Kafle and Kanan(2017)}]{kafle2017analysis}
Kushal Kafle and Christopher Kanan. 2017.
\newblock An analysis of visual question answering algorithms.
\newblock In \emph{Proceedings of the IEEE International Conference on Computer
  Vision}, pages 1965--1973.

\bibitem[{Kervadec et~al.(2021)Kervadec, Antipov, Baccouche, and
  Wolf}]{kervadec2021roses}
Corentin Kervadec, Grigory Antipov, Moez Baccouche, and Christian Wolf. 2021.
\newblock Roses are red, violets are blue... but should vqa expect them to?
\newblock In \emph{Proceedings of the IEEE/CVF Conference on Computer Vision
  and Pattern Recognition}, pages 2776--2785.

\bibitem[{Kim et~al.(2018)Kim, Jun, and Zhang}]{kim2018bilinear}
Jin-Hwa Kim, Jaehyun Jun, and Byoung-Tak Zhang. 2018.
\newblock Bilinear attention networks.
\newblock \emph{Advances in Neural Information Processing Systems}, 31.

\bibitem[{Liang et~al.(2021)Liang, Hu, and Zhu}]{liang2021lpf}
Zujie Liang, Haifeng Hu, and Jiaying Zhu. 2021.
\newblock Lpf: A language-prior feedback objective function for de-biased
  visual question answering.
\newblock \emph{arXiv preprint arXiv:2105.14300}.

\bibitem[{Liang et~al.(2020)Liang, Jiang, Hu, and Zhu}]{liang2020learning}
Zujie Liang, Weitao Jiang, Haifeng Hu, and Jiaying Zhu. 2020.
\newblock Learning to contrast the counterfactual samples for robust visual
  question answering.
\newblock In \emph{Proceedings of the 2020 Conference on Empirical Methods in
  Natural Language Processing (EMNLP)}, pages 3285--3292.

\bibitem[{Mahabadi and Henderson(2019)}]{mahabadi2019simple}
Rabeeh~Karimi Mahabadi and James Henderson. 2019.
\newblock Simple but effective techniques to reduce biases.
\newblock \emph{arXiv preprint arXiv:1909.06321}, 9.

\bibitem[{Niu et~al.(2021)Niu, Tang, Zhang, Lu, Hua, and
  Wen}]{niu2021counterfactual}
Yulei Niu, Kaihua Tang, Hanwang Zhang, Zhiwu Lu, Xian-Sheng Hua, and Ji-Rong
  Wen. 2021.
\newblock Counterfactual vqa: A cause-effect look at language bias.
\newblock In \emph{Proceedings of the IEEE/CVF Conference on Computer Vision
  and Pattern Recognition}, pages 12700--12710.

\bibitem[{Oord et~al.(2018)Oord, Li, and Vinyals}]{oord2018representation}
Aaron van~den Oord, Yazhe Li, and Oriol Vinyals. 2018.
\newblock Representation learning with contrastive predictive coding.
\newblock \emph{arXiv preprint arXiv:1807.03748}.

\bibitem[{Pennington et~al.(2014)Pennington, Socher, and
  Manning}]{pennington2014glove}
Jeffrey Pennington, Richard Socher, and Christopher~D Manning. 2014.
\newblock Glove: Global vectors for word representation.
\newblock In \emph{Proceedings of the 2014 conference on empirical methods in
  natural language processing (EMNLP)}, pages 1532--1543.

\bibitem[{Ramakrishnan et~al.(2018)Ramakrishnan, Agrawal, and
  Lee}]{ramakrishnan2018overcoming}
Sainandan Ramakrishnan, Aishwarya Agrawal, and Stefan Lee. 2018.
\newblock Overcoming language priors in visual question answering with
  adversarial regularization.
\newblock \emph{arXiv preprint arXiv:1810.03649}.

\bibitem[{Ren et~al.(2015)Ren, He, Girshick, and Sun}]{ren2015faster}
Shaoqing Ren, Kaiming He, Ross Girshick, and Jian Sun. 2015.
\newblock Faster r-cnn: Towards real-time object detection with region proposal
  networks.
\newblock \emph{Advances in neural information processing systems}, 28:91--99.

\bibitem[{Robinson et~al.(2020)Robinson, Chuang, Sra, and
  Jegelka}]{robinson2020contrastive}
Joshua Robinson, Ching-Yao Chuang, Suvrit Sra, and Stefanie Jegelka. 2020.
\newblock Contrastive learning with hard negative samples.
\newblock \emph{arXiv preprint arXiv:2010.04592}.

\bibitem[{Selvaraju et~al.(2019)Selvaraju, Lee, Shen, Jin, Ghosh, Heck, Batra,
  and Parikh}]{selvaraju2019taking}
Ramprasaath~R Selvaraju, Stefan Lee, Yilin Shen, Hongxia Jin, Shalini Ghosh,
  Larry Heck, Dhruv Batra, and Devi Parikh. 2019.
\newblock Taking a hint: Leveraging explanations to make vision and language
  models more grounded.
\newblock In \emph{Proceedings of the IEEE/CVF International Conference on
  Computer Vision}, pages 2591--2600.

\bibitem[{Si et~al.(2021)Si, Lin, Zheng, Fu, and Wang}]{si2021check}
Qingyi Si, Zheng Lin, Mingyu Zheng, Peng Fu, and Weiping Wang. 2021.
\newblock Check it again: Progressive visual question answering via visual
  entailment.
\newblock \emph{arXiv preprint arXiv:2106.04605}.

\bibitem[{Tan and Bansal(2019)}]{tan2019lxmert}
Hao Tan and Mohit Bansal. 2019.
\newblock Lxmert: Learning cross-modality encoder representations from
  transformers.
\newblock \emph{arXiv preprint arXiv:1908.07490}.

\bibitem[{Teney et~al.(2020)Teney, Abbasnejad, Kafle, Shrestha, Kanan, and Van
  Den~Hengel}]{teney2020value}
Damien Teney, Ehsan Abbasnejad, Kushal Kafle, Robik Shrestha, Christopher
  Kanan, and Anton Van Den~Hengel. 2020.
\newblock On the value of out-of-distribution testing: An example of goodhart's
  law.
\newblock \emph{Advances in Neural Information Processing Systems},
  33:407--417.

\bibitem[{Wu and Mooney(2019)}]{wu2019self}
Jialin Wu and Raymond Mooney. 2019.
\newblock Self-critical reasoning for robust visual question answering.
\newblock \emph{Advances in Neural Information Processing Systems}, 32.

\bibitem[{Zhu et~al.(2020)Zhu, Mao, Liu, Zhang, Wang, and
  Zhang}]{zhu2020overcoming}
Xi~Zhu, Zhendong Mao, Chunxiao Liu, Peng Zhang, Bin Wang, and Yongdong Zhang.
  2020.
\newblock Overcoming language priors with self-supervised learning for visual
  question answering.
\newblock \emph{arXiv preprint arXiv:2012.11528}.

\end{thebibliography}

\appendix

\section{More Details of the Proposed Method}
\begin{table*}[t]
\centering
\resizebox{1.0\linewidth}{!}{
\begin{tabular}{l|p{185pt}|p{180pt}|p{150pt}} 
 \hline
    Type&original & \textbf{\emph{Shuffle}} &  \textbf{\emph{Removal}} \\ \hline
  Y/N &Is this indoors or outside ? & Is ? indoors outside or this   & indoors or outside ?  \\ 
  Y/N & Are these buildings new ? &   new these buildings ? Are & buildings new ? \\ 
  Y/N &Does this person eat healthily ? &this ? person healthily eat Does &  person eat healthily ? \\
  Num & How many people will be dining ?& ? be many people How will dining & people will be dining ?\\
  Num &How many small zebra are there ?& there zebra small ? are How many& small zebra are there ? \\
  Other &What is the smallest kid holding ? & the is smallest What ? holding kid & smallest kid holding ?\\
  Other & Who is on the screen ?&Who screen ? the is on& on the screen ?\\
  Other &What are people wearing on their heads ?&their are wearing ? on people heads What& people wearing on their heads ?\\
   
  Other &What animals are walking on the road ?&  road the are on What animals ? walking&animals are walking on the road ?\\
  Other & What color is the food inside the bowl ? & the color the food What is bowl inside ? & food inside the bowl ? \\
  
  \hline 
\end{tabular}
}
\caption{More examples of two types of positive samples.}
\label{exampls}
\end{table*}

\begin{table}
    \centering

    \resizebox{1.0\linewidth}{!}{
    \begin{tabular}{l|cllll}
    \hline
        Type & n($C_{qtype}$) & m($Z_C$) & m($W_C$)\%& m($P_C$)\% & m($Z_C^{unb}$) \\ \hline
        Y/N & 28 & 209  & 92.60 & 18.52 & 39  \\ \hline
        Num & 4 & 156  & 56.84 & 11.37 & 19  \\ \hline
        Other& 33 & 836 & 3.76& 0.75 & 10  \\ \hline

    \end{tabular}

    }
\caption{The statistics about the question type (e.g., Y/N) and the corresponding unbiased samples with the setting of $\beta$=20\%. For all question categories (e.g, what color) in each question type, ($C_{qtype}$) represents the number of them; m($Z_C$) represents the mean value of their label space size; m($W_C$) represents the mean value of their correction factors which are used to dynamically adjust $\beta$; m($P_C$) represents the mean value of their unbiased answer proportions after being adjusted; m($Z_C^{unb}$) represents the mean value of their unbiased answer number.}
\label{statistics}

\end{table}

\subsection{Discussion about the positive samples.}\label{app1.1}

We give more examples of \textbf{\emph{Shuffling}} and  \textbf{\emph{Removal}} positive questions in Tab. \ref{exampls}. We can see that the intention of the `Y/N' questions can still be inferred from the \textbf{\emph{Removal}} questions. By contrast, the intention of the \textbf{\emph{Removal}} questions for non-`Y/N' questions is ambiguous. This attests to the rationality of the proposed \textbf{\emph{SR}} strategy, which treats `Y/N' and non-`Y/N' questions differently.


Although the positive samples could cause some confusion/ambiguity, it may not impact our method too much, because: 1) In MBSS, the model only makes prediction on the original samples during training, and thus it does not directly associate the answers with the positive questions, which are only used in contrastive learning. 2) \textbf{\emph{Shuffling}} could change the original questions to a conflicting meanings, e.g., , `How many bananas are next to the apples?' and `How many apples are next to the bananas?'. However, such special cases are very rare. For a question whose length is 7\footnote{The average length of questions in the training set is 7.14}, the probability of shuffling to a conflicting meaning is $\frac{1}{7!}$. In most cases, the \textbf{\emph{Shuffling}} just eliminates the sequential information of the questions, but basically conveys the same meaning. 3) In terms of \textbf{\emph{Removal}}, we only construct this kind of positive questions for the `Y/N' questions, which does not change the intended meaning of the original question as discussed in the above paragraph. 4) Additionally, the proposed unbiased sample selection module prevents the potential noise in positive questions from affecting the unbiased samples, which are beneficial to OOD generalization. 


\subsection{Unbiased sample statistics. }
\label{app1.2}
To further investigate how the unbiased-sample-selection algorithm treats different types of questions , i.e. `Y/N', `Num' and `Other' questions, we roughly divide all the question categories into  the three types according their semantics, and then do some statistical analysis about the question types and the corresponding unbiased samples. We set the initial unbiased answer proportion (hyper-parameter) $\beta$ = 20\%. As the detail statistics shown in Tab. \ref{statistics}, we find that: 1) the `Other' questions have the largest answer space while the `Num' questions have the smallest one. Counter-intuitively, the `Y/N' questions also have a relatively large number of candidate answers. For example, `red' is also annotated as the answer to the question `Is this flower red?'. However, this rarely happens compared with the answer `yes'.  2) The proposed correction factor $W_C$ is close to 1 when the question is a `Y/N' question and the  $W_C$ is close to 0 when the question is a `Other' question. Correspondingly, the adjusted unbiased answer proportion $P_C$ is close to $\beta$ for `Y/N' questions while it is relative smaller for `Other' questions. This is consistent with the phenomenon that most ground truth of `Y/N' questions concentrate on much fewer answers (e.g., `Yes') than that of `Other' questions. 

\section{More Experimental Setups}\label{app2}

\begin{table}[t]
\centering
\resizebox{.95\columnwidth}{!}{
\begin{tabular}{l|lllll} 
 \hline
    Model & $Epo$ & $\alpha$ &  $\beta$  & $Lr$ & $N'$\\ \hline
    BAN+Ours & 25 & 1 & 0.5 & 1e-4&- \\
    UpDn+Ours & 60 & 1 & 0.6 & 1e-4 &-\\ 
    LXM+Ours & 40 & 1 & 0.2 & 5e-6/5e-5&- \\ 
    LMH+Ours & 60 & 0.18 & 0.5 & 1e-4&-\\ 
    LXM+LMH+Ours & 40 & 0.18& 0.2 & 5e-6/5e-5 & - \\
    U-SAR+Ours  & 10 & 0.18 & 0.5 & 1e-5&2,20 / 2,2 \\
    SAR+Ours & 10 & 0.18 & 0.5 & 1e-5& 2,20/ 2,20\\  \hline
\end{tabular}
}
\caption{The detailed hyper-parameter settings of our methods. The $Epo$ represents the number of training epochs. $Lr$ represents the initial learning rate of Adam optimizer on VQA-CP v2/VQA v2. $N'$, is a SAR-specific hyper-parameter, represents the number of candidate answers for yesno, non-yesno questions during test on VQA-CP v2/VQA v2.}
\label{table_para}
\end{table}

\begin{table}[t]
\centering
\resizebox{.95\columnwidth}{!}{
\begin{tabular}{l|llp{70pt}} 
 \hline
    Model & Param. & Training Time  & Infrastructure \\ \hline
    UpDn+Ours & 36M  & 0.38h/epo & TITAN RTX 24GB GPU\\ \hline
    LXM+Ours & 213M  & 1.73h/epo  & 2 x TITAN RTX 24GB GPUs\\ \hline
\end{tabular}
}
\caption{The details of computational experiments of our methods based on UpDn and LXM.}
\label{compute}
\end{table}
\subsection{Implementation details.}
Following existing works, we use the Faster R-CNN \citep{ren2015faster} to extract fixed 36 objects feature embeddings with 2048 dimensions for each image. All the questions are trimmed or padded to 14 words. For the UpDn backbone model, we apply a single-layer GRU to encode the word embeddings( initialized with Glove \citep{pennington2014glove}) of the question into a 1280-dimensional question embeddings. We follow \citep{zhu2020overcoming} and adopt a multi-step learning rate that halves every 5 epochs after 10 epochs. For the LXMERT backbone, we use the tokenizer of LXMERT to segment each input question into words. We adopt the cosine learning rate decay following the warmup in the first 5 epochs. We train the models with batch size of 128. The detailed hyper-parameter settings of our methods in the main results are shown in Tab. \ref{table_para}. The details of computational experiments of our method based on UpDn and LXMERT are shown in Tab. \ref{compute}. We keep the same random seed during training and testing for \emph{\textbf{Shuffling}} method. As the change of seed has little effect on each method, following most of previous works, we also report the results with a single run. 

\subsection{Positive sample construction for SAR.}
SAR \citep{si2021check} is a two-stage framework: it first selects the most relevant candidate answers, and then combines the question and each candidate answer to produce \textit{dense captions}, and finally, reranks the dense captions based on visual entailment. They design two ways to construct the dense captions, including 1) replacing the question category prefix with answer and 2) concatenating question and answer directly. To apply MMBS to SAR, we construct the positive dense captions for the rerank stage. Specifically, we directly use the first kind of captions as \textbf{\emph{S}} positive captions, because the question category prefix has already been removed. For the second kind of captions, we randomly shuffle the words to construct the \textbf{\emph{R}} positive captions. The input dense caption during training and test are the second kind of captions. Following \citet{si2021check}, we set the number of candidate answers for training to 20. During test, we set the number of the candidate answers to $N'$ shown in Tab. \ref{table_para}.

\begin{table}\footnotesize
    \centering
    \resizebox{0.88\linewidth}{!}{
    \begin{tabular}{l|llll}
    \hline
        Method & All & Y/N & Num & Other \\ \hline
        UpDn & 41.06 & 43.13 & 13.71 & 47.48 \\
        UpDn\emph{+orig.} & 41.39 & 42.23 & 13.7 & 48.54 \\ 
        UpDn\emph{+rand-SR} & 44.21 & 51.19  & \textbf{15.05} & 48.56  \\ 
        UpDn\emph{+SR} & \textbf{47.62} & \textbf{62.72} & 13.92 & \textbf{48.95} \\  \hline

        LXM & 47.19 & 50.55 & 24.06 & 51.77 \\ 
        LXM\emph{+orig.} & 48.14 & 51.25 & 25.63 & \textbf{52.69} \\
        LXM\emph{+rand-SR }& 51.07 & 62.22 & \textbf{29.68} & 51.09 \\ 
        LXM\emph{+SR }& \textbf{55.26} & \textbf{77.13} & 27.33 & 51.47 \\ \hline
        LMH & 52.01 & 72.58 & 31.12 & 46.97 \\ 
        LMH\emph{+orig. }& 55.25 & 74.84 & \textbf{41.11} & 48.87 \\ 
        LMH\emph{+rand-SR }& \textbf{55.50} & 75.36 & 35.67 & \textbf{50.54} \\ 
        LMH\emph{+SR }& 55.41 & \textbf{76.50} & \textbf37.20 & 49.35 \\ \hline

    \end{tabular}
    }
\caption{Results on VQA-CP v2 for validating the effectiveness of \textbf{\emph{SR}} strategy. The models here do not contain the unbiased sample selection module. }
\label{AblationStudy2}
\vspace{-0.2cm}
\end{table}
\section{More Experiments and Analysis}\label{app3}

\subsection{Further validation of the effectiveness of \emph{\textbf{SR}} strategy. }
To better validate the effectiveness of \emph{\textbf{SR}} strategy, we also evaluate the model performance  directly using the original sample as positive sample ( \emph{+orig.}), or randomly adopting one of  \textbf{\emph{S}} and  \textbf{\emph{R}} as positive sample ( \emph{+rand-SR}) for each sample.  
We can observe from Tab. \ref{AblationStudy2} that: 1) \emph{+orig.} constantly outperforms the backbone models because the contrastive learning itself is helpful for learning a better feature representation. 
2) It is worth noting that when we apply \emph{+orig.} on LMH, the performance improvement is much more obvious. This is because ensemble-based methods have relieved the language priors to some extent at the cost of almost entirely attenuating the positive information from the biased samples. Our method makes up for this drawback and forces the model to pay attention again to this information by minimizing contrastive learning loss which does not cause superficial correlations, unlike the normal VQA loss. This can also explain that the performance of \emph{+orig.}, \emph{+rand-SR} and \emph{+SR} is similar based on the ensemble-based methods. 3) For UpDn and LXM: a) \emph{+rand-SR} outperforms \emph{+orig.} considerably, which demonstrates that the design of positive samples by excluding the correlations between the question category and answer benefits MMBS in overcoming language priors; b) Compared with \emph{+rand-SR}, \emph{+SR} achieves prominent performance boost on `Y/N' questions, and slightly improves the performance or maintains competitive performance on the other two types of questions, which attests to the soundness of the motivation of strategy \textbf{\emph{SR}}. 

\end{document}